\documentclass[10pt,twocolumn,letterpaper]{article}

\usepackage{cvpr}              

\usepackage{graphicx}
\usepackage{amsmath}
\usepackage{amssymb}
\usepackage{booktabs}
\usepackage[dvipsnames]{xcolor}
\usepackage{colortbl}

\usepackage[accsupp]{axessibility}  

\usepackage{multirow}
\usepackage{multicol}

\usepackage{color, colortbl}
\definecolor{Gray}{gray}{0.9}
\definecolor{LightCyan}{rgb}{0.88,1,1}
\usepackage[first=0,last=9]{lcg}

\definecolor{LightBlue}{rgb}{0.90,0.95,1}

\usepackage[font=small,labelfont=bf,tableposition=top]{caption}
\usepackage{blindtext}
\usepackage{cuted}

\usepackage[pagebackref,breaklinks,colorlinks]{hyperref}

\def\eg{\emph{e.g}\onedot} 
\def\ie{\emph{i.e}\onedot}

\usepackage[capitalize]{cleveref}
\crefname{section}{Sec.}{Secs.}
\Crefname{section}{Section}{Sections}
\Crefname{table}{Table}{Tables}
\crefname{table}{Tab.}{Tabs.}

\def\cvprPaperID{9384} 
\def\confName{CVPR}
\def\confYear{2023}

\title{OSRE: Object-to-Spot Rotation Estimation for Bike Parking Assessment}

\begin{document}
\author{Saghir Alfasly$^{1,2}$ \:\:\: Zaid Al-huda$^{3}$\:\:\: Saifullah Bello$^{1}$ \:\:\: Ahmed Elazab$^{4}$ \:\:\: Jian Lu$^{1,}$\thanks{Corresponding author}  \:\:\: Chen Xu$^{1,2}$ \\
		\small $^{1}$Shenzhen Key Laboratory of Advanced Machine Learning and Applications, Shenzhen University, China\\
		\small $^{2}$Guangdong Key Laboratory of Intelligent Information Processing, China\\
			 \small $^{3}$ School of Computing and Artificial Intelligent,	Southwest Jiaotong University, China\\
			 \small $^{4}$ School of Biomedical Engineering, Shenzhen University, China\\
		{\tt\small \{saghiralfasly, jianlu, yuruzou\}@szu.edu.cn}, {\tt\small chenxuszu@sina.com}
	}	
\maketitle

  \begin{strip}
	\centering\noindent
	\includegraphics[trim=0 0 0 0, width=17.5cm,clip, keepaspectratio]{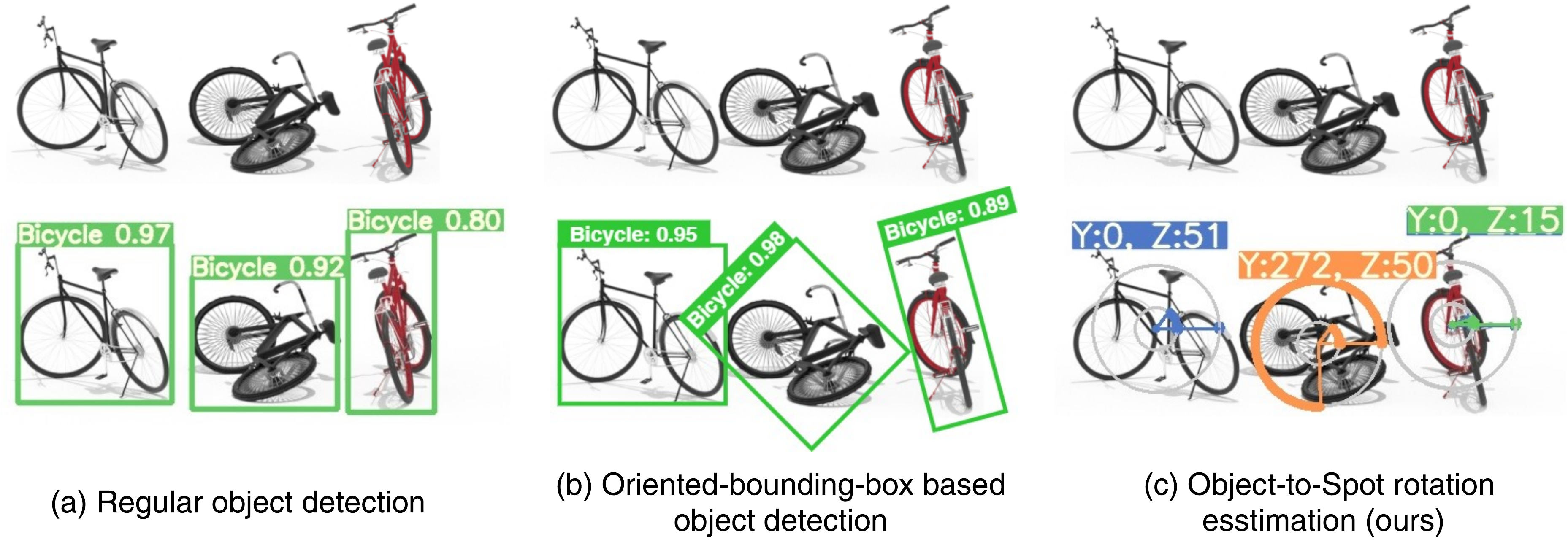}\vspace*{-3mm}
	\captionof{figure}{\textbf{Object-to-Spot Rotation Estimation.} Our proposed object-to-spot rotation estimation method provides two rotation angle predictions in two axes $y$ and $z$. Rotation in $y$ axis represents the bicycle leaning/falling left or right. When it is fallen down to the left side, it represents the angle $90^\circ$ or $\pi/2$ in radians, whereas falling down to the right size is representing the $270^\circ$ or $3\pi /2$. Similarly, rotation in $z$ axis represents the bike direction in its standing pose. \ie, (C) shows the rotated bike in $z$ axis ({\tt rotated} class) in \textcolor{blue}{blue} color, the bikes lean left/right ({\tt fallen} class) are shown in \textcolor{orange}{orange} color, and the well-parked bikes ({\tt parked} class) are plotted in \textcolor{Green}{green}  color. We visualize $z$ rotation in the inner circle and $y$ is visualized in the outer circle. Fig.\ref{fig:RotationVisualized} illustrates the rotation visualization in more details.}
	\label{fig:overview}
\end{strip}\vspace*{-2mm}

\medskip
\vspace*{-5mm}	
\begin{abstract}\vspace*{-3mm}	
Current deep models provide remarkable object detection in terms of object classification and localization. However, estimating object rotation with respect to other visual objects in the visual context of an input image still lacks deep studies due to the unavailability of object datasets with rotation annotations.

This paper tackles these two challenges to solve the rotation estimation of a parked bike with respect to its parking area. First, we leverage the power of 3D graphics to build a camera-agnostic well-annotated Synthetic Bike Rotation Dataset (SynthBRSet). Then, we propose an object-to-spot rotation estimator (OSRE) by extending the object detection task to further regress the bike rotations in two axes. Since our model is purely trained on synthetic data, we adopt image smoothing techniques when deploying it on real-world images. The proposed OSRE is evaluated on synthetic and real-world data providing promising results. Our data and code are available at \href{https://saghiralfasly.github.io/OSRE/}{https://saghiralfasly.github.io/OSRE/}.\vspace*{-1mm}

\end{abstract}
		
	\section{Introduction} \vspace*{-1mm}
	\label{sec:intro}\
The rapid advancement in computational and communication technology, remarkably, benefits intelligent surveillance systems and smart cities applications. Among them, automatic traffic congestion analysis, bike-sharing and smart parking space management systems which substantially become in demand in recent years \cite{smartCity,smartCity2}.

\begin{figure}
	\centering
	\includegraphics[trim=0 0 0 0, width=8.3cm,clip, keepaspectratio]{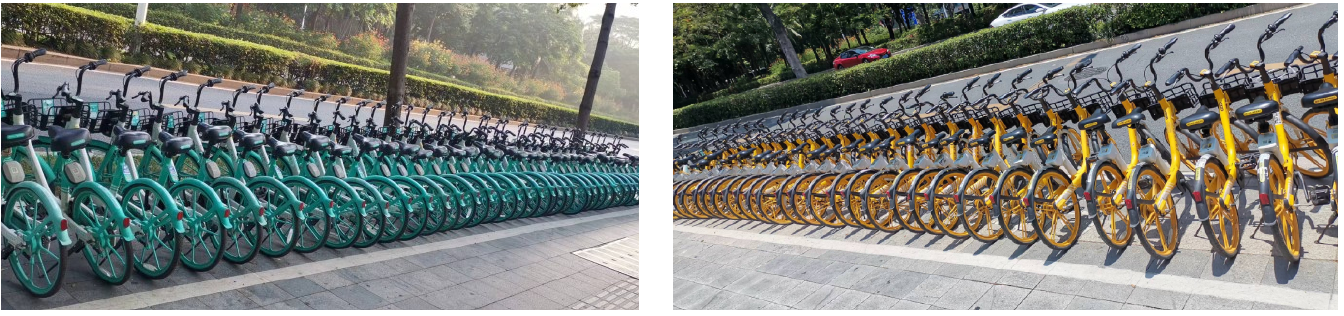}
	\vspace*{-3mm}
	\caption{\textbf{Bike-Sharing Companies.} Bicycles organized manually by bike-sharing companies. The current bike-sharing systems include dock and dockless systems. Dockless systems feature smart algorithms to locate bikes. Thus, the problem of {\tt\small \textbf{where}} to park the bike is effectively solved, whereas, {\tt\small \textbf{How}} to park the bike is still an open question and challenging problem to tackle. The cost in terms of time and labor spent by such companies to keep bikes well organized and distributed motivated us to build a computer vision deep model that can effectively assess the bike parking and thus it can be used as a primitive step for further alert or alarm for park them appropriately.}
	\label{fig:realBikeWellParked}
\end{figure}

Despite the overall great service provided by bicycle-sharing systems, the growing number of users has posed challenges to integrated management such as keeping bicycles well organized and distributed, in particular dockless systems. Nevertheless, current dockless bicycle-sharing systems effectively manage bike parking using location tracking technology such as GPS, and thus, these systems drive users {\tt \textbf{where}} to park bikes. Unfortunately, these systems fail to lead users {\tt \textbf{how}} to park bikes. Hence, several bicycle-sharing companies tend re-distribute and organize bikes manually as shown in Fig. \ref{fig:realBikeWellParked}, which questions such method's efficiency.

Generally, surveillance-based systems can be equipped with vision-based deep models to assess bike parking. However, current object detection methods do only tackle object localization and categorization which cannot be explicitly applied to the bike parking problem. In addition, existing regular \cite{dai1605object,girshick2014rich,girshick2015fast,ren2015faster,he2017mask,lin2017feature,cai2018cascade,hu2018relation,tian2019fcos,redmon2016you,redmon2017yolo9000,redmon2018yolov3,bochkovskiy2020yolov4,wang2022yolov7,liu2016ssd,lin2017focal} and oriented-bounding-box object detection methods \cite{
	ma2018arbitrary,jiang2018r,yang2019scrdet,liao2018textboxes,xie2021oriented,cheng2022dual,cheng2022anchor} likely ignore object-to-object rotations and only focus on fitting the bounding box to the best tight object boundaries as shown in Fig. \ref{fig:overview}. Thus, estimating object rotation with respect to other visual objects in image visual context has not been tackled yet. More specifically, the orientation of a parked bike with respect to its parking area is a key challenging task for bike parking assessment. To tackle this problem, a large visual bike parking dataset is required with each bike rotation annotation which is, to the best of our knowledge, not publicly available.

\noindent\textbf{Contributions.}
To this end, we present a bike parking assessment method by tackling both challenges: the data unavailability and the lack of deep models for object rotation estimation. \textit{First}:: We leverage the power of 3D graphics to build a Synthetic Bike Rotation Dataset (SynthBRSet) that is camera-agnostic and well-annotated. Although it is a challenging scheme to rely only on the visual synthetic data for training deep models to tackle such real-world problems, it may provide exceptional and sufficient data due to two main reasons. On one hand, bike object rotation estimation does not rely mainly on fine-grained details such as textures, materials, and colors. Instead, it focuses on higher semantic visual properties which are easy to synthetically generate with 3D graphics. On the other hand, the difficulty of manually collecting real-world balanced datasets with various bike poses, rotations, types, and density makes the choice of 3D graphics engine more likely to be effective. \textit{Second}: We propose a deep object-to-spot (bike to the parking region) rotation estimator (OSRE). Our object rotation estimator performs two tasks. It predicts and localizes bikes with their parking classes (\ie, {\tt parked}, {\tt rotated}, and {\tt fallen}). Additionally, it regresses the bike rotation with respect to the parking area in two axes $y$ and $z$. Note that parking areas are semantically derived from the input image rather than explicitly defined or annotated. Moreover, as our model is trained purely on synthetic data, we adopt image smoothing techniques when deploying it on real-world images as shown in the right part of Fig. \ref{fig:samplePrediction}. The proposed OSRE is evaluated on synthetic and real-world data providing promising results. \textbf{Overall, the key contributions of this work are as follows}:
	\vspace*{-1mm}
	\begin{itemize}
		\item Leveraging the power of 3D graphics, we created Synthetic Bike Rotation Dataset (SynthBRSet) that features diversity, generalization, and balance for effectively training deep bike rotation estimation models.
	\vspace*{-2mm}
	\item We propose an object-to-spot rotation estimation model, which is, to the best of our knowledge, the first attempt to build an end-to-end object-to-spot rotation estimation (OSRE) model that regresses the rotations of objects in two axes $y$ and $z$. 
	\vspace*{-2mm}
	\item When deploying OSRE to run on real-world images, we adopt image smoothing/filtering techniques to bridge the gap between the synthetic and real data, which provides a remarkable performance boost.
		\vspace*{-2mm}
	\end{itemize}
	
\section{Related Work}
\noindent\textbf{3D Graphics and Visual Synthetic Data.}
Deep learning models rely on massive amounts of data for effective training. In order to have better generalization, these data need to be sampled from diverse environments. Recently, the use of synthetic data for training is becoming popular \cite{sergy.synthetic.data, Weinzaepfel:VirtualGallery:CVPR2019, SunSPWGSTY22, YanZRLD22}. Synthetic data allows the creation of a 3D environment with the objects of interest appropriately placed to reflect the real world. This provides the benefit of manually labeling a few objects of interest and using them to create a big repository of data augmented with variations such as backgrounds, noise, objects' color and orientation. Synthetic data can also be obtained from other existing sources like video games \cite{Johnson-Roberson17, RichterVRK16, RichterHK17, Guo_2022_CVPR}. Synthetic data is used in various applications in computer vision. For example, the popular datasets for 3D recognition\cite{modelnet} and parts segmentation\cite{shapenetparts} are based on synthetic CAD models. Synthetic data have also been used for optical flow estimation\cite{ButlerWSB12, MayerIHFCDB16, MayerIFHCDB18}, depth estimation\cite{MayerIHFCDB16, Krahenbuhl18}, and visual odometry\cite{ZhangRFS16, HandaNAD12, RichterHK17}. In more high-level applications, synthetic data have been used in semantic segmentation \cite{HuCHHS19, HandaPBSC15a, KimWLK20, Krahenbuhl18, Varol0MMBLS17}; object tracking \cite{7780839, FabbriBMCGOCLC21, SunASSMS20}, human pose estimation \cite{FabbriLCAC20, ShottonGFSCFMKCKB13}, and video representations \cite{Guo_2022_CVPR}. In object detection, which is closely related to this work, synthetic data are used in \cite{HowardZCKWWAA17, AmatoCFGM19, MarinVGL10}. Howard et al. \cite{HowardZCKWWAA17} used synthetic data on pre-trained object detectors with frozen lower layers to obtain results close to training on a large real dataset. Photorealistic 3D videos were used to extract synthetic images in \cite{AmatoCFGM19} for pedestrian detection. Scene-specific pedestrian detection was proposed by Hattori [31] that was purely trained using synthetic objects of interest superimposed on real scenes. In this work, we used purely synthetic data to train the model; moreover, we validated the proposed model on real data.

	\begin{figure}
	\centering
	\includegraphics[trim=0 0 0 0, width=8.3cm,clip, keepaspectratio]{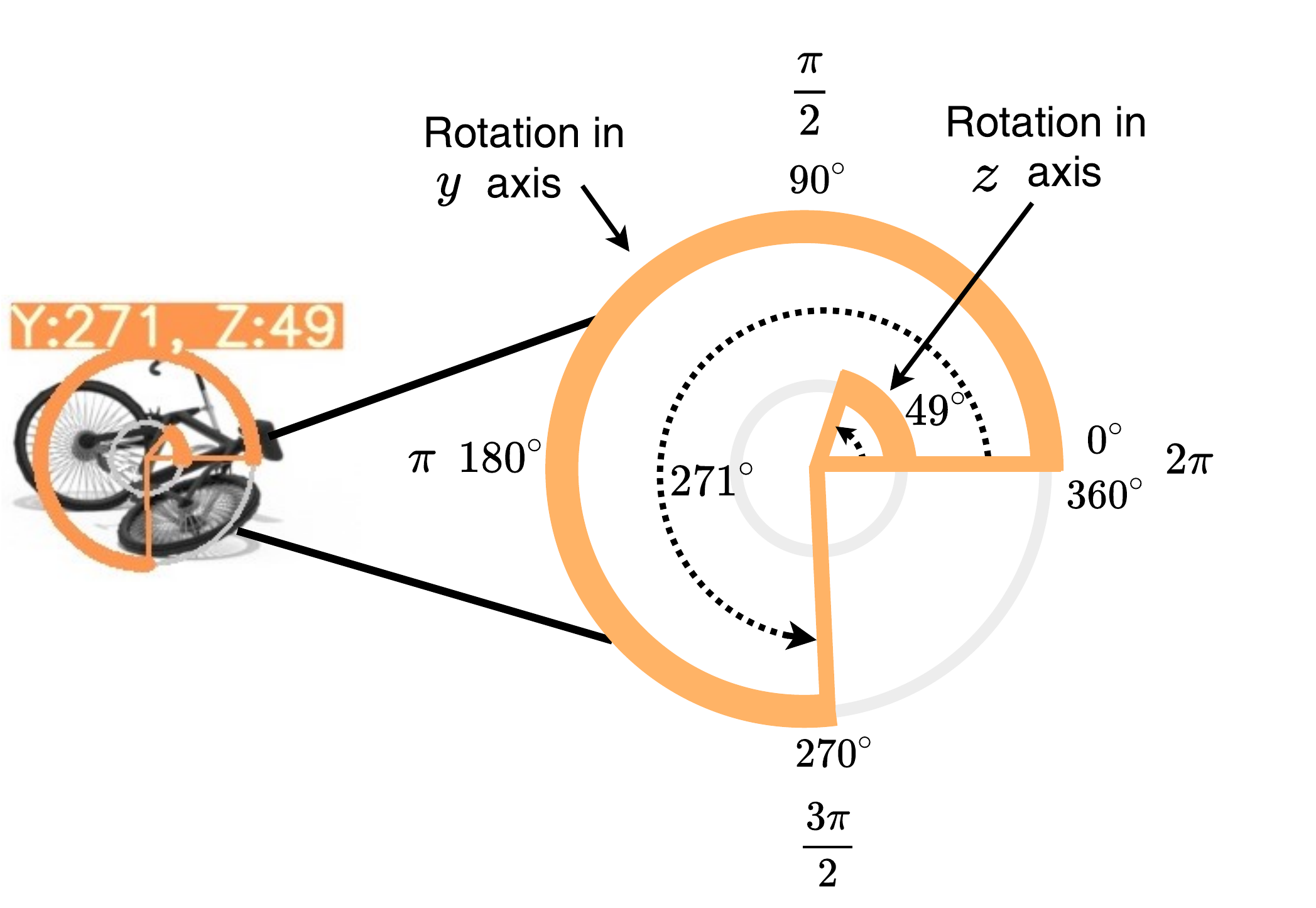}
	\caption{\textbf{Rotation Visualization.} We represent rotation in two axes, the outer angle represents the rotation of the object (\eg, bike) in $y$ axis, whereas the inner angle represents the object rotation in $z$ axis. We assume the bike well parked is not rotated in either axes. Rotating the bike in $y$ axis leads to lending the bike and/or fall it down in the ground. The rotated bike in $y$ axis may and may not be associated with rotation in $z$ axis at the same time. However, bikes can be rotated only in $z$ axis which indicates that it is standing and parked but in an inaccurate direction.}
	\label{fig:RotationVisualized}
\end{figure}

\begin{figure*}
	\centering
	\includegraphics[trim=0 0 0 0, width=17.5cm,clip, keepaspectratio]{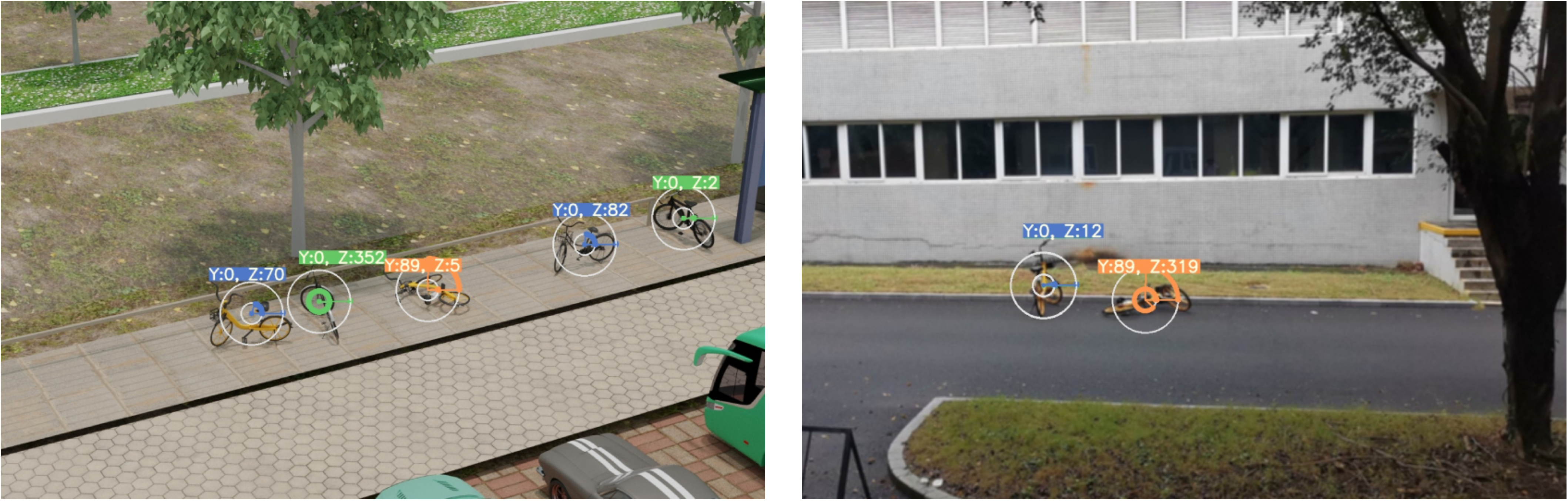}
	\caption{\textbf{Bike Rotation Estimation on Synthetic and Real Data}. Left is a bike rotation estimation on a synthetic image which shows accurate prediction. On the other hand, a prediction sample on real image is shown on the right. Similarly, the estimated rotation is clearly accurate as it represents the visual bike pose. Our rotation visualization is depicted in more detail in Figs. \ref{fig:RotationVisualized} and \ref{fig:bikeRotationYZ}}
	\label{fig:samplePrediction}
\end{figure*}

\noindent\textbf{Object Detection.}
In many computer vision applications, object detection is an essential task for other subsequent tasks, receiving significant attention from both academia and industry \cite{chen2021points,peng2022context,zhou2021instant}, \ie, automatic transportation, unmanned aerial vehicles, and monitoring. The advancement of deep neural networks (NN) has led to three main types of horizontal object detection methods: two-stage, one-stage, and keypoint-based detectors. In the two-stage detectors \cite{dai1605object,girshick2014rich,girshick2015fast,ren2015faster,he2017mask,lin2017feature,cai2018cascade,hu2018relation}, the dense object proposals are generated first, followed by prediction and refinement of the final bounding box boundaries. A number of high-performance two-stage detectors have been developed, including R-CNN \cite{girshick2014rich}, Fast R-CNN \cite{girshick2015fast}, Faster R-CNN \cite{ren2015faster}, and Mask R-CNN \cite{he2017mask}.   

Although the two-stage detectors continue to attract much attention, another research area focuses on the design and development of efficient one-stage detectors \cite{tian2019fcos,redmon2016you,redmon2017yolo9000,redmon2018yolov3,bochkovskiy2020yolov4,wang2022yolov7,liu2016ssd,lin2017focal, alfaslyzoom} due to their much simpler and cleaner construction. There are several single-stage detectors, including YOLO \cite{redmon2016you} and its variants (YOLO9000 \cite{redmon2017yolo9000}, YOLOv3 \cite{redmon2018yolov3}, YOLOv4 \cite{bochkovskiy2020yolov4}, YOLOv5, YOLOR, and YOLOv7 \cite{wang2021your,wang2022yolov7}), SSD \cite{liu2016ssd}, and RetinaNet \cite{lin2017focal}.  On the other hand, keypoint-based detectors \cite{duan2019centernet,law2018cornernet,zhang2021adaptive,yang2019reppoints} (e.g., CenterNet \cite{duan2019centernet} and Cornernet \cite{law2018cornernet}) group objects into final bounding boxes by exploring informative key points (such as points and corners).

However, horizontal object detection detectors (Fig. \ref{fig:overview}.(a)) are unable to rotate the bounding boxes to fit the detected objects. Thus, new kinds of object detectors are proposed (Fig. \ref{fig:overview}-(b)), are proposed for fitting bounding boxes with the detected objects. In Rotation Region Proposal Network (RRPN) \cite{ma2018arbitrary}, the rotated region of interest (ROI) is obtained from the rotated anchor, and its depth feature is extracted. Using RPN architecture, a rotation candidate box has been introduced for the first time to detect scenes in any direction. Rotational Region CNN (R2CNN) \cite{jiang2018r} is a text detector that detects both rotated and horizontal bounding boxes simultaneously. SCRDet \cite{yang2019scrdet} enhances R2CNN by incorporating a fusion of features and spatial and channel attention mechanisms. Despite the number of proposals increasing, TextBoxes++ \cite{liao2018textboxes} is able to accommodate the narrowness of the text with a long convolution kernel. Recently, an oriented Region Proposal Network (oriented RPN) \cite{xie2021oriented} was proposed to generate high-quality oriented proposals on a near-cost-free basis. Using dual-aligned oriented detectors (DODets) \cite{cheng2022dual}, rotated proposals can be efficiently generated and feature misalignment can be reduced between classification and localization. In contrast to the anchor-based approach, Cheng et al. \cite{cheng2022anchor} introduced an innovative Anchor-free Oriented Proposal Generator (AOPG) that eliminates all horizontal boxes from the network. The development of rotated object localization has been greatly facilitated by these excellent methods.

Aligning anchor boxes/regions of interest (RoIs) and convolution features is an essential step for both one-stage and two-stage detectors. It is challenging to achieve accurate detection using detectors that rely on misaligned features. Fixed-length features are extracted from RoIs by a RoI operator (such RoIPooling \cite{girshick2015fast}, RoIAlign \cite{he2017mask}, and Deformable RoIPooling \cite{dai2017deformable}) to approximatively locate objects in two-stage detectors. Following the partition of a RoI into sub-regions, RoIPooling max-pools each sub-region into the respective output grid cell. Due to the quantization of the floating-number boundary of a RoI into integer form, RoIPooling leads to a misalignment between the RoI and the feature. In RoIAlign, bilinear interpolation is used to avoid quantization of RoIPooling, which has the advantage of improving localization performance. With Deformable RoIPooling, features can be selected according to their context based on offsets added to each subregion of a RoI.

	\textbf{Our work differs from previous methods in the following aspects.}
	\textbf{\textit{First}}, our object-to-spot rotation estimator is remarkably different than regular object detectors \cite{dai1605object,girshick2014rich,girshick2015fast,ren2015faster,he2017mask,lin2017feature,cai2018cascade,hu2018relation,tian2019fcos,redmon2016you,redmon2017yolo9000,redmon2018yolov3,bochkovskiy2020yolov4,wang2022yolov7,liu2016ssd,lin2017focal}. Object detection models tend to predict object class and its location. It optimizes the object localization ability (\ie, \textit{where} the object located in the spatial space). Along with the aforementioned predictions, our rotation estimator further predict the object rotations with respect to the spot of the parking area derived from the image semantic context. Thus, it tends to estimate \textit{how} the bikes are located.
	\textbf{\textit{Second}}, Oriented bounding box based detectors \cite{
		ma2018arbitrary,jiang2018r,yang2019scrdet,liao2018textboxes,xie2021oriented,cheng2022dual,cheng2022anchor} tend to predict the object bounding box rotation with respect to the overall input frame (\ie, object rotation with respect to the capturing camera view). However, our method predicts the object rotation with respect to another visual object in the same input frame. Additionally, our method is agnostic to the camera viewing angle as it predicts the same rotations of a bike from different camera angles.
	\textbf{\textit{Finally}}, enriching the training data with additional synthetic data greatly boosts the performance of deep models. However, our model is trained purely on the generated synthetic data due to the difficulty of manually annotating object rotation. Nevertheless, it shows promising results when it runs on real data.

	\begin{figure*}
	\centering
	\includegraphics[trim=0 0 0 0, width=17.0cm,clip, keepaspectratio]{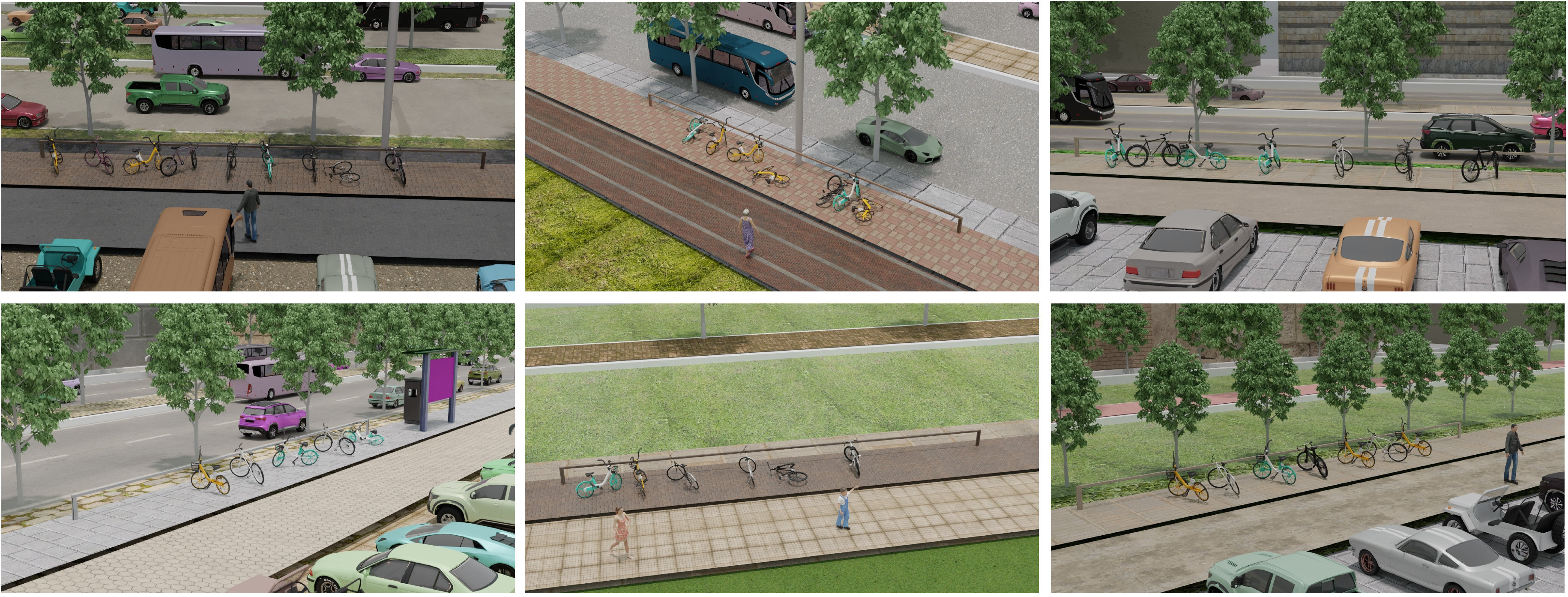}
	\vspace*{-2mm}
	\caption{\textbf{Generated Image Samples}. Synthetic Bike Rotation Dataset (SynthBRSet) contains a wide variety in lighting, objects' colors, objects' poses, camera view, and scenes. More samples are shown in the appendix.}
	\label{fig:imageSamples}
\end{figure*}

\begin{figure}
	\centering
	\includegraphics[trim=0 0 0 0, width=9.0cm,clip, keepaspectratio]{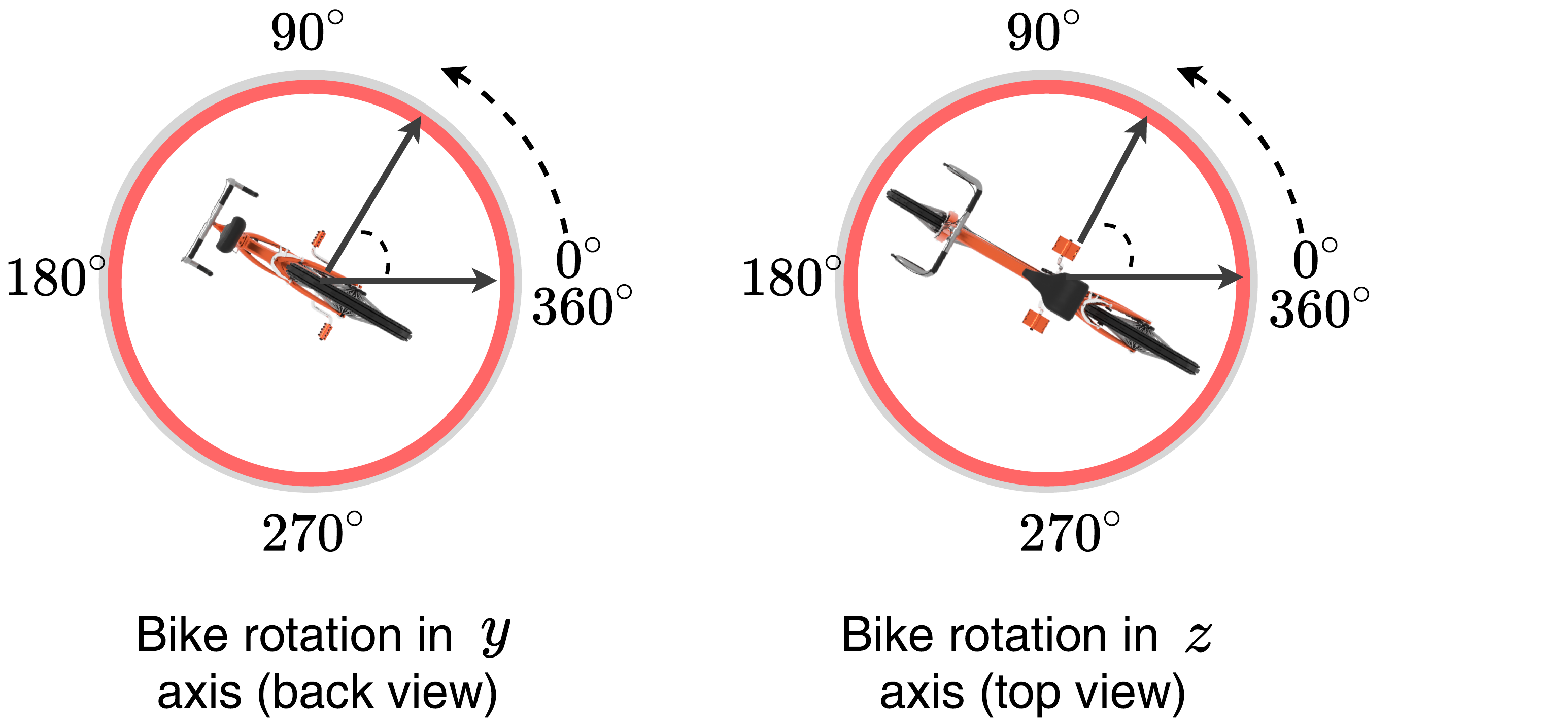}
	\caption{\textbf{Bike Rotation Axes.} Bike rotation is represented in $y$ axis as shown in the left part, which can be observed from the bike's back view. On the other hand, bike rotation in $z$ axis is shown in the right part which is visualized as the top view of the bikes. }
	\label{fig:bikeRotationYZ}
\end{figure}
	\section{Proposed Bike Rotation Estimator}
	\label{sec:method} 
	\subsection{Synthetic Bike Parking Dataset (SynthBRSet)}
Generating a real dataset for object rotation is a huge challenge due to the difficulty of collecting the visual object annotation by physically recording their rotation with respect to a specific spatial region. For example, collecting a dataset for bike rotation estimation with respect to the base parking area requires manual recording of the physical measurement of each bike's rotation angle relative to the spot. As it is inefficient and hugely laborious to annotate it manually, it is hard to obtain accurate rotation reading or even closer to real rotation reading. This challenge is may be doubled when we target to train deep models which require large datasets. Thus, it is of significant motivation to leverage the power of 3D computer graphics to generate such datasets with accurate annotations in terms of both bike localization and its rotations.

We leveraged Blender \cite{blender} to build an algorithm that generates a synthetic visual bike parking dataset. The dataset consists of large variations in terms of parking space, lighting conditions, backgrounds, material textures, and colors. It automatically and randomly localizes a variant number of bikes with different visual and physical properties in a parking spot. Then, it calculates their rotations in $y$ and $z$ with respect to the parking spot as shown in Fig. \ref{fig:bikeRotationYZ} and visualized in Fig. \ref{fig:RotationVisualized}. 
It is worth mentioning that the spot parking area is not visually annotated or considered since we trained our model to implicitly estimate its location and rotation. Thus, we avoid deep models from overfitting to specific visual parking spot properties. Overall, we can summarize the properties of the proposed synthetic bike parking algorithm as follows: 
	
	\noindent\textbf{Bike Variation.}
We model several bikes as well as use several models to increase the bike variations in terms of bike structures and color which in turn increases the diversity of the generated dataset. We used several various bike models which can be further extended for improving the dataset generalization. 

\noindent\textbf{Rotation Variation.}
The main aim to use 3D graphics software is to generate accurate rotation annotations. Along with each bike bounding box and class, we annotate its rotation with respect to the parking spot area. We acquire the bike rotations in two axes: $y$ axis represents the horizontal bike rotation (\ie, back view in Fig. \ref{fig:bikeRotationYZ}) and $z$ axis represents the vertical bike rotation (\ie, top view in Fig. \ref{fig:bikeRotationYZ}).

\noindent\textbf{Objects, Backgrounds Variation.}
Although our aim is to generate a bike dataset with high variation in terms of bike type, size, color, texture, and size, we enrich the dataset by randomly changing several objects with various properties. For example, vehicles with variant sizes and crowding density. Additionally, we randomly add pedestrians with different sizes, gender, ages, and clothes colors, standing or walking in different directions.

\noindent\textbf{Lighting, Material, Texture and Color Variation.} We apply variant lighting, color, materials, and textures to increase the dataset's visual diversity which in turn improves the generalization of the trained model in this data. We designed our algorithm to randomly change the color of the vehicles and bikes. It also randomly changes the ground/road textures and backgrounds. For example, we use the grass with different textures as we use different road backgrounds for the road spaces as shown in Fig. \ref{fig:imageSamples}.

\noindent\textbf{Camera Variation.} One of the most challenges in training deep vision models is to make the view generalization which is a difficult requirement to meet in most image- and video-based datasets. Thus, generating 3D visual data provides unconditional control in generating several samples of a specific image from different viewpoints. Our dataset is generated with a wide range of camera angles that mimic the potential real surveillance camera locations. This variation leads to building a generalized model that is agnostic to the camera angles.
	
	 \begin{figure*}
	 	\centering
	 	\includegraphics[trim=0 0 0 0, width=17.0cm,clip, keepaspectratio]{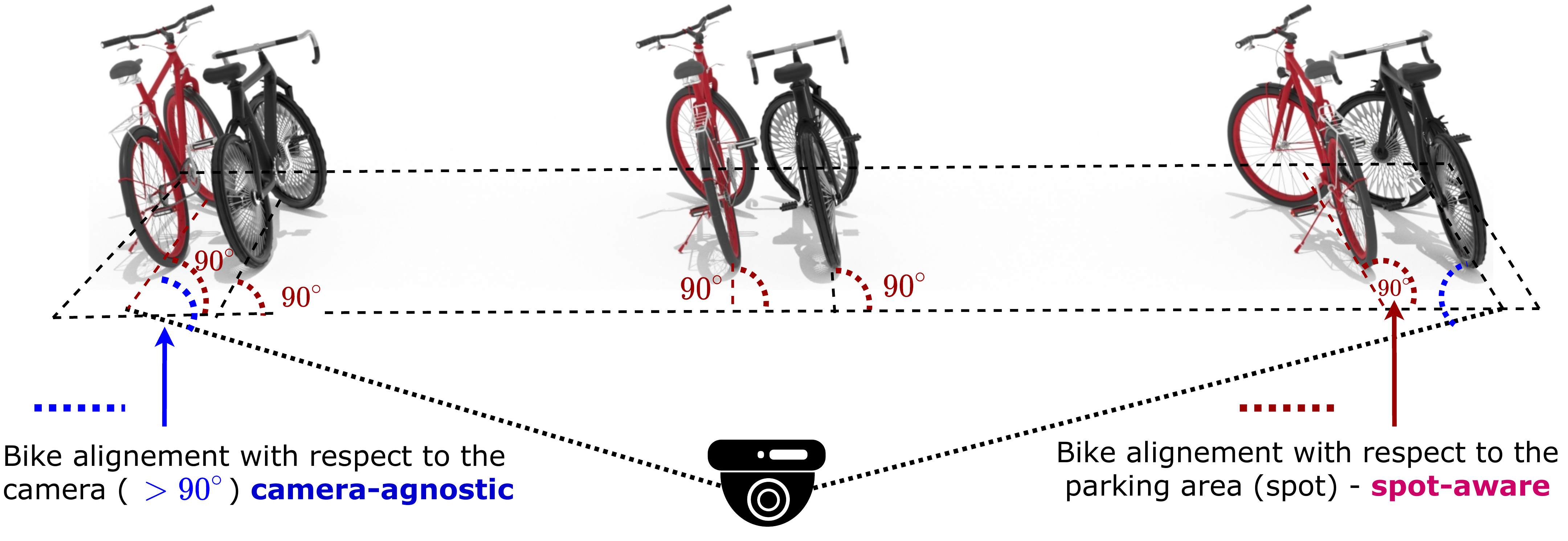}
	 	\vspace*{-3mm}
	 	\caption{\textbf{Camera-Agnostic Spot-Aware Rotation Illustration}. All these bikes are accurately aligned and well-parked. However, they have different views which are commonly challenging in computer vision to recognize and identify objects from a single view. This motivated us to leverage the power of 3D graphics to generate accurate spot-aware rotation annotations with a large variety of camera views.}
	 	\label{fig:IMD}
	 \end{figure*}\vspace*{-2mm} 
 
	\subsection{Object-to-Spot Rotation Estimator}
Since we obtained a large bike dataset with accurate rotation annotations, here, we propose a deep object-to-spot (bike to the parking region) rotation estimation. Our proposed model features two prediction tasks. It predicts and localizes the bike parking from the three classes (\ie, parked, rotated, and fallen). Additionally, it regresses the bike rotation in two axes $y$ and $z$ with respect to the parking area. Note that the parking area on the input image is implicitly derived from the input image rather than defined explicitly in annotations. In other words, in the training phase, we feed the model with the bike rotation annotation without their target spot (\ie, parking area) annotations. This forces the model to learn to estimate the parking region and then predict bike rotation with respect to it. We followed this scheme to prevent model overfitting as the model will get fit easily to the parking area rotation and location.

\begin{figure}
	\centering
	\includegraphics[trim=0 0 0 0, width=8.0cm,clip, keepaspectratio]{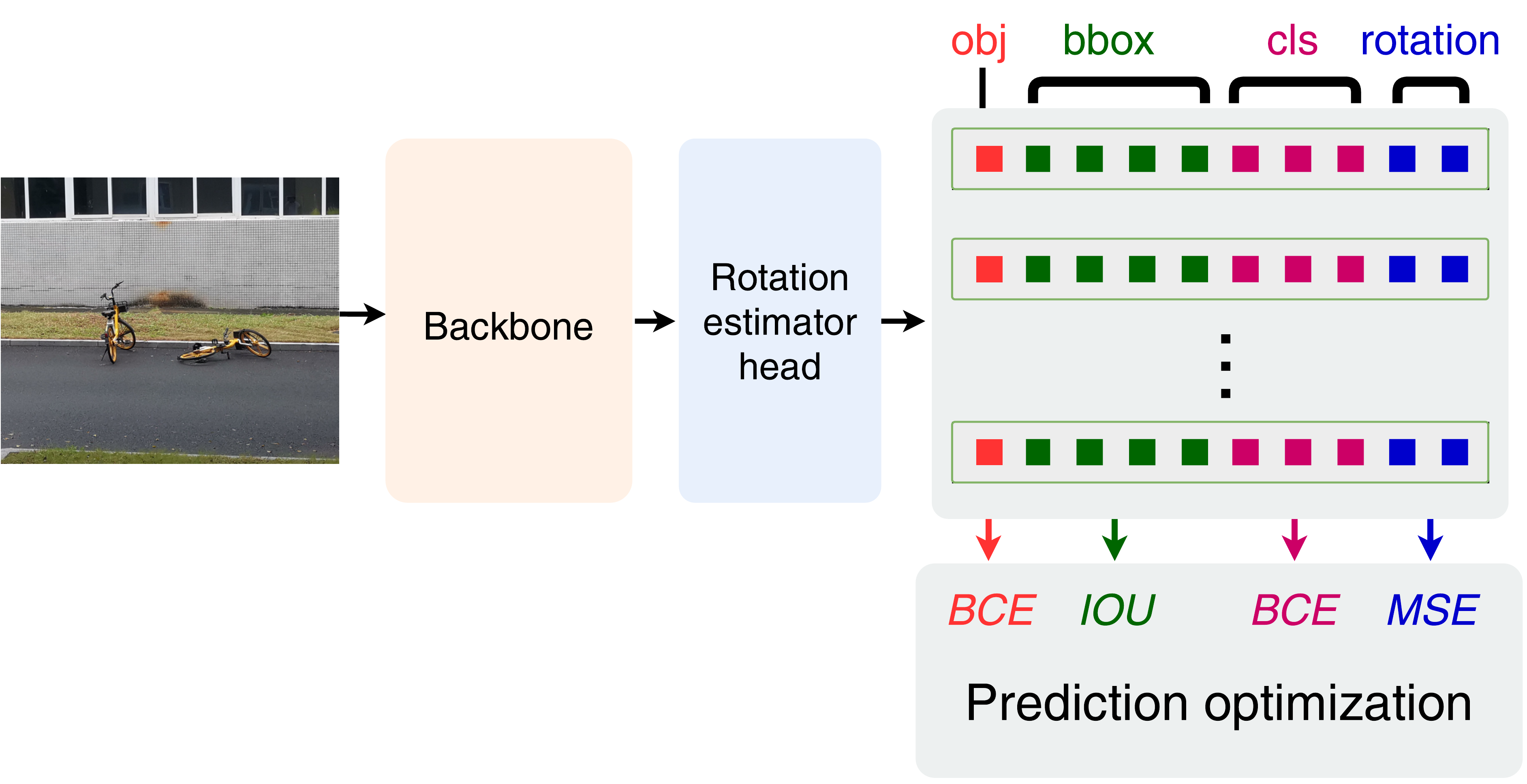}
	\vspace*{-3mm}
	\caption{\textbf{Rotation Estimator Training.} We adapt object detection baselines with an extension on the prediction head for which it predicts the rotation angles in $y$ and $z$ axes. Similarly to Yolo and most single-shot object detection models, bounding boxes are optimized with intersection over union (IOU), whereas the object presence and class predictions are optimized with binary cross-entropy (BCE). For optimizing the rotations, we adopt mean square error (MSE) in a regression scheme.}
	\label{fig:rotationModel}
\end{figure}

	As our model features most object detection properties such as localizing all potential bikes in the input image and then classifying all bikes into three classes:  {\tt\small parked} which refers to the well-parked bike status;  {\tt\small rotated} refers to the status when the bike is parked (\ie, standing), but in a different direction which translated in a rotation in $z$ axis.  {\tt\small fallen} refers to the bike state when it is fallen down reflecting the rotation in axis $y$ as in Fig.\ref{fig:bikeRotationYZ}. We adapt several feature extractors of Yolo versions \cite{redmon2016you,redmon2017yolo9000,redmon2018yolov3, bochkovskiy2020yolov4, wang2022yolov7}. The prediction head is designed to further predicts rotation angles. Thus, for each predicted bounding box, we regress the potential bike rotation in both axes. Overall, we adapt most Yolo7 components \cite{wang2022yolov7}.

The proposed mode regresses the bike rotation as a logistic regression problem that is optimized with mean squared error (MSE). As shown in Fig. \ref{fig:rotationModel}, we adapt Yolo loss function to optimize the object presence as logistic regression with binary cross-entropy. Regressed bounding box predictions are optimized with IOU, whereas the bike class prediction is optimized with a logistic classifier with binary cross-entropy.
	\vspace*{-2mm}
	
\subsection{Real-world Image Smoothing}
The main reason for the low performance of a deep model trained on synthetic visual data is the low-level feature gap between the synthetic images and real images, where synthetic images have smoother textures and materials. Real-world images are more diverse and empowered by huge variations in textures and surfaces. Hence, we attempt to bridge the gap between synthetic and real-world images by applying image smoothing/filtering techniques including 2D convolution, low-pass filtering, gaussian filtering, median filtering, and bilateral filtering. A sample of an image smoothed with 2D convolution is shown in the right part of Fig. \ref{fig:samplePrediction}. We further discuss these techniques' settings in the next section.
	
\begin{table*}
	\fontsize{8}{11}\selectfont
	\centering
	
	\caption{Generated Synthetic Dataset. It consists of two main sets. The first row contains the challenging set with a more dense crowd as compared to the second set. Along with the number of bike objects, we summarize the number of each parking status class.}    \vspace*{-4mm}
	\begin{tabular}{ccccccccc}
		\bottomrule
		\multirow{2}{1.0cm}{\centering Data}  & \multirow{2}{1.0cm}{\centering Camera} & \multirow{2}{1.0cm}{\centering Images} &\multirow{2}{*}{\centering Bike Objects} & \multicolumn{3}{c}{Classes} & \multirow{2}{1.0cm}{Train}  & \multirow{2}{1.0cm}{ Test}\\
		\cline{5-7} 
		&&&&Parked & Rotated & Fallen&&\\\hline
		Challenging& free&$ 4,440 $&$ 42,247 $&$ 22,421 $&$ 11,847 $&$ 7,979 $&$ 4,012 $ &$ 429 $ \\\hline
		\multirow{3}{*}{\centering Regular Set}&free &$11,843$&$  71,366  $&$ 26,789 $&$ 26,589 $&$ 17,988 $&$ 10,659 $& $ 1,184 $ \\
		&vertical-free &$2,000$ &$ 12,953 $&$ 4,798 $&$ 4,861 $&$ 3,294 $&$ 1,793 $& $ 207 $ \\
		&restricted &$2,000$ &$ 13,015 $&$ 4,950 $&$ 4,889 $&$ 3,176 $&$ 1,790 $& $ 210 $ \\
		\hline \rowcolor{LightBlue}\hline
		Total &-&$20,284$ &$ 139,581 $&$ 58,958 $&$ 48,186 $&$ 32,437 $&$ 18,234 $& $ 2,050 $ \\
		\toprule
	\end{tabular}
	\label{tab:dataset}
\end{table*}

\section{Experimental Results}
\subsection{Implementation Details.}
We built our object-to-spot rotation estimation model on Yolo7 \cite{wang2022yolov7}. Where we used similar settings and augmentation techniques. We built our synthetic data generator as a non-parametric 3D graphics based procedure in Blender \cite{blender}. We designed several bike models and other objects. Several material assets are collected from a real-world in real bike parking scenarios\footnote{Our {\tt.blend} file and python script to generate a synthetic bike dataset are attached in the supplementary file. We will make them available online as well. \href{https://drive.google.com/drive/folders/1GK5oK9EeIoCch1fZlZxrnMjWp1aT3Ks9?usp=sharing}{}}. Additionally, we use several open-source modelled objects such as vehicles and trees. We run our generating procedure on machine with $2$ $\times$ RTX $ 2080 $ Ti to generate $ 360 \times 640 $ output images. The approximate rendering time per image is $ 42 $ second. Regarding the object-to-spot rotation estimator, we built it on Yolo7 \cite{wang2022yolov7}. Hence, we ensure we build our model on the state-of-the-art object detector that can be run in real-time for industrial real-world applications. We followed most Yolo training recipes and evaluating settings along with our rotation evaluating and optimization modules. Our training and evaluation tasks are conducted on $8$ $\times$ RTX $3090$. Unless otherwise specified, we run all ablated experiments on the same machine to ensure the consistency of performance readings. All our trained model use $ 640 \times 640 $ input dimension. Further implementation details are provided in the supplementary file. \vspace*{-1mm}

	\subsection{Experiments Setup}
\noindent\textbf{Datasets.} The generated SynthBRSet contains $ 139,581 $ bikes with varying pose and parking rotation angles in $ 20,284 $ images. All bike objects are categorized into three main parking statuses based on their rotation with respect to the parking region:  {\tt\small parked},  {\tt\small rotated}, and  {\tt\small fallen}. Since we have full data generating control, we generated several variants. For example, we generated the first set as a challenging set that involves less intra-bike distance and bikes between $ 3 $ and $ 20 $, which makes the deep model learning more challenging. This set contains $ 4,440 $ image. The second set involves a regular dataset with a bike crowd count between $ 3 $ and $ 15 $. This set contains $13,843$ images. Among this set, we have one subset of $ 2,000 $ images with a fixed camera, and one subset of $ 2,000 $ with a horizontally-constrained camera and vertically free. Table \ref{tab:dataset} shows the dataset details, where each dataset is split into ($ \%9:\%1 $) train-test sets. Additionally, to test the proposed method in real data, we collected and annotated $175$ bike objects with varying parking poses. This real subset contains $134$ {\tt\small parked}, $28$ {\tt\small rotated}, and $18$ {\tt\small fallen} bikes in total $ 33 $ images. It is worth mentioning that collecting and annotating the bike rotations in real-world scenarios is a quite challenging task as we are required to measure each angle of $y$ and $z$ axes\footnote{Details on the process of collecting real-world bike set with annotated rotations are further given in the supplementary file}.

\begin{figure*}
	\centering
	\includegraphics[trim=0 0 0 0, width=17.5cm,clip, keepaspectratio]{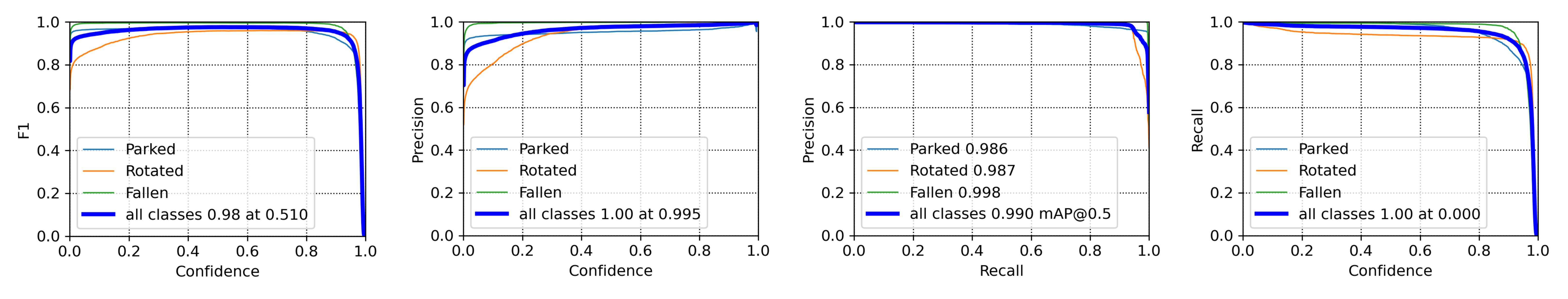}
	\vspace*{-7mm}
	\caption{\textbf{Performance of OSRE on Synthetic Testing Data}. Performance of OSRE in terms of F1/confidence, precision/confidence, precision/recall, and recall/confidence.}
	\label{fig:results}
\end{figure*}\vspace*{-2mm} 

\noindent\textbf{Evaluation Metrics.} Along with the mean square error used for bike rotation assessment, we further evaluate our proposed model with respect to object detection and localization. Thus, we report the performance in terms of  precision/recall, F1/confidence, and mean average precision (mAP). We further visualize the output results by plotting the aforementioned metrics.

\begin{figure}
	\centering
	\includegraphics[trim=0 0 0 0, width=8.0cm,clip, keepaspectratio]{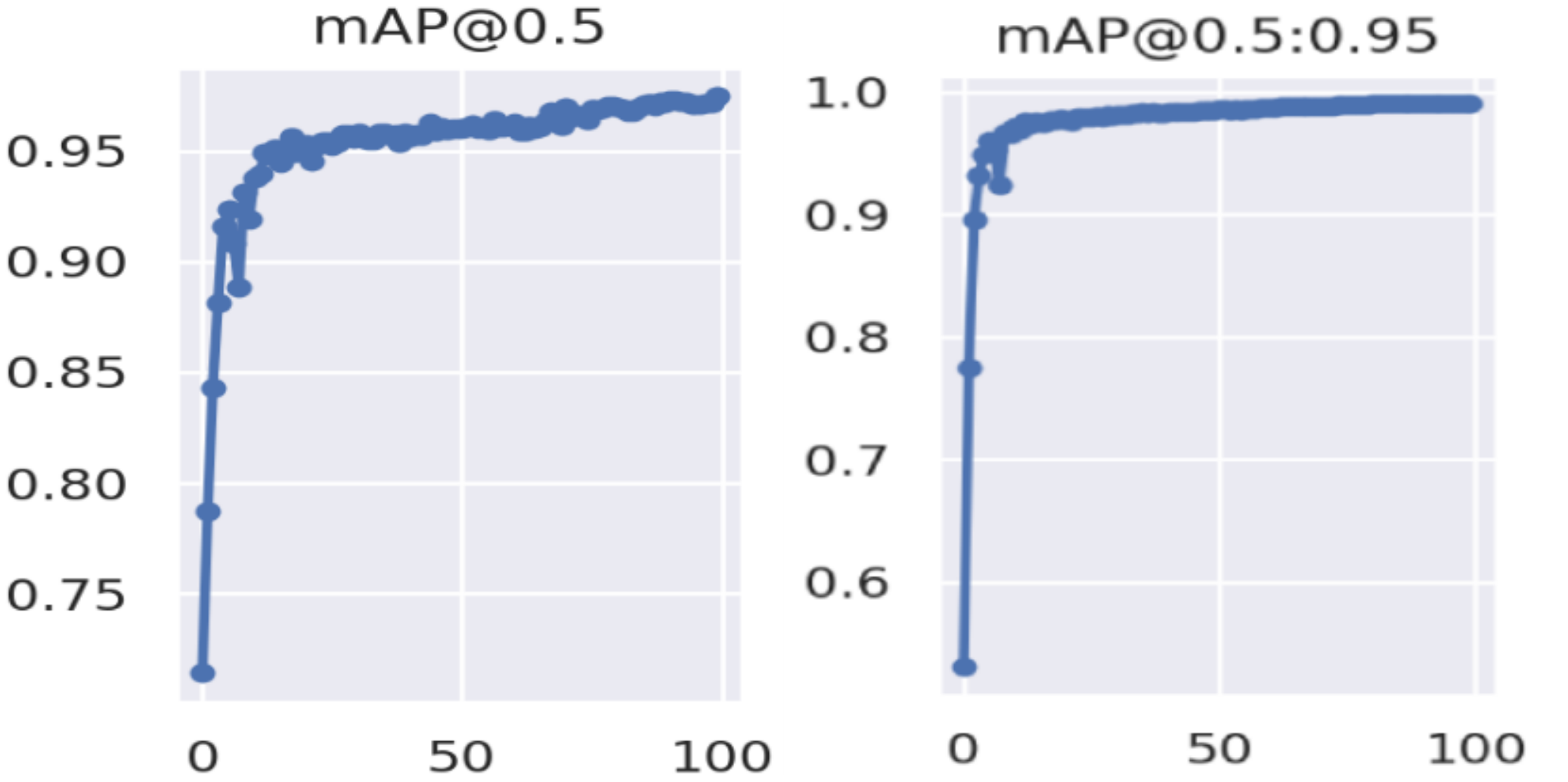}
	\vspace*{-3mm}
	\caption{\textbf{OSRE for Bike Detection.} Performance of the trained OSRE on the synthetic data. The model is trained on the entire data and evaluated on its testing set.}
	\label{fig:resultsmAP}
\end{figure}
\subsection{Results}
	In this section, we conduct a set of experiments to evaluate the performance of the proposed bike-to-spot rotation estimator on several state-of-the-art object detection baselines. Moreover, we ablate the proposed model in terms of the training data and loss weighting.\footnote{More results are reported in the supplementary file}
	
	\noindent\textbf{OSRE on State-of-the-art Baselines.}
We adapt several baselines as a backbone for our model for bike rotation estimation. Here, we trained Yolo version variants including Yolov7 \cite{wang2022yolov7}, Yolov3 \cite{redmon2018yolov3}, Yolov4 \cite{bochkovskiy2020yolov4}, Yolor \cite{wang2021your} with various backbones including Darknet \cite{redmon2018yolov3}, darknet53, ResNeXt \cite{ResnextXie2016}, and ResNet \cite{resnet} associated with CSPNet \cite{cspnet} to improve the model structure. Table \ref{tab:OSRE} reports the performance results of the proposed OSRE on top of various backbones. Remarkably, it provides the best performance when we adapt Yolov7-X with $ 0.083 $ and $ 0.028 $ MSE on the synthetic and real data, respectively. This reflects Yolov7 superiority in regular object detection tasks. Figs. \ref{fig:results} and \ref{fig:resultsmAP} show results in bike detection metrics. 

\begin{table}
	\fontsize{8}{10}\selectfont
	\centering   
	\caption{Performance of the proposed OSRE on top of several object detection baselines. The models are trained on half SynthBPSet excluding the challenging set and evaluated on SynthBPSet and the real sets.} \vspace*{-4mm}
	\begin{tabular}{lcccc}
		\bottomrule
 \multirow{2}{*}{\centering Model} & \multicolumn{2}{c}{ Detect (AP)}& \multicolumn{2}{c}{ Rotate (MSE)} \\
		
		\cline{2-3} \cline{4-5} 
		& Synth. & Real & Synth. & Real \\
		\toprule
		OSRE-YOLOv3-CSP \cite{redmon2018yolov3}&$ 93.4 $ &$ 38.7 $&$ 0.084 $& $ 0.034 $\\
		OSRE-YOLOv4-CSP \cite{bochkovskiy2020yolov4}&$ 93.1 $ &$ 30.1 $&$ 0.082 $ & $ 0.040 $\\
		OSRE-YOLOR-CSP \cite{wang2021your}&$ 93.7 $ &$ 45.0 $&$ 0.082 $& $ 0.049 $ \\
		OSRE-YOLOv7 \cite{wang2022yolov7}&$ 93.7 $ &$ 38.2 $&$ 0.081 $& $ 0.043 $\\\rowcolor{LightBlue}
		OSRE-YOLOv7-X \cite{wang2022yolov7}&$ 93.8 $ &$ 29.9 $& \textbf{0.083}   &  \textbf{0.028} \\
		\toprule
	\end{tabular}
	\label{tab:OSRE}
\end{table}

%

\noindent\textbf{Ablation Study.}
Since four of our OSRE elements are required to be optimized including the rotations ($ Ryz $), object presence ($ obj $), bounding boxes ($ bbox $), and the bike classes ($ cls $), we studied the impact of weighting our model loss $\gamma_{Ryz} $. We kept the weights $\gamma_{obj}=0.7 $, $\gamma_{cls}=0.3 $, $\gamma_{bbox}=0.05 $, adapt several baselines as a backbone for our model for bike rotation estimation. Table \ref{tab:loss} reports the performance of the proposed OSRE with various rotation loss weight. Notably, it provides the best performance when we assign $0.05$ weight to the rotation element. This is reasonable since the main aim of the training model is to pay more attention to the rotation rather than the bounding box fit. As results show, the best reported rotation estimation is $0.028$ MSE and detection $0.51$ mAP on the real-world testing set. Finally, we evaluate the impact of various smoothing and filtering techniques on the quality of rotation estimation performance. Table \ref{tab:smoothing} reports the performance reports showing the 2D Convolution with $5\times5$ kernel size and the low-pass filter of size $3\times3$ the best rotation estimation results with $ 0.028 $ MSE.

\begin{table}
	\fontsize{8}{10}\selectfont
	\centering   
	\caption{Performance of the proposed model with varying rotation loss weights. Model variants are trained on the entire dataset} \vspace*{-4mm}
	\begin{tabular}{lcccccccc}
		\bottomrule
		\multirow{2}{*}{\centering $\gamma_{obj} $} & \multirow{2}{*}{\centering $\gamma_{ bbox} $} & \multirow{2}{*}{\centering $\gamma_{cls} $} & \multirow{2}{*}{\centering $\gamma_{Ryz} $} & \multicolumn{2}{c}{ Detect (AP)}& \multicolumn{2}{c}{ Rotate (MSE)} \\
\cline{5-6} \cline{7-8}
&&&& Synth & Real& Synth & Real \\
\toprule
		0.7&0.05&0.3& \textbf{0.3} & $97.2 $ &$ 40.6 $&$ 0.0779 $&$ 0.037 $\\
		
		 0.7&0.05&0.3&\textbf{0.1} & $ 97.4 $ &$ 47.5 $&$ 0.0779 $&$ 0.030 $\\ \rowcolor{LightBlue}
		 0.7&0.05&0.3& \textbf{0.05} & $96.1 $ & \textbf{51.0} &$ 0.077 $& \textbf{0.028} \\
		 0.7&0.05&0.3& \textbf{0.02} &$  97.4 $&$ 49.7 $&$ 0.077 $&$ 0.031 $\\
		\toprule
	\end{tabular}
	\label{tab:loss}
\end{table}
\begin{table}
	\fontsize{8}{10}\selectfont
	\centering   
	\caption{Impact of real-world image smoothing. OSRE is trained with $\gamma_{Ryz}$=$0.05$ and tested on the real dataset.} \vspace*{-4mm}
	\begin{tabular}{lccc}
		\bottomrule
		Smoothing/Filtering & Kernel/parameters& Rotation (MSE) \\ 
		\toprule
		None & - &$ 0.047 $\\ \rowcolor{LightBlue}
		2D Convolution & $1/25 \times (5\times5)$ & \textbf{0.028} \\ \rowcolor{LightBlue}
		Low-pass filter & $1/9 \times (3\times3)$ &\textbf{0.028} \\
		Gaussian Filtering & $(5\times5), 0$ &$ 0044 $\\
		Median Filtering & $(5\times5)$ &$ 0.047 $\\
		Bilateral Filtering & $(5\times5)$ &$ 0.046 $\\
		\toprule
	\end{tabular}
	\label{tab:smoothing}
\end{table}
	\subsection{Discussion and Limitations}
Based on the performance of the proposed object-to-spot rotation estimator and the reported results, it is worth mentioning a few observations and recommendations for further investigation.
\textbf{\textit{Firstly}}, leveraging the power of 3D graphics to generate photorealistic data significantly benefit difficult deep-learning-based models, especially those in which some annotation is quite challenging to collect manually. 
\textbf{\textit{Secondly}}, Although, our proposed method is mainly built upon single-shot object detectors (\ie, YOLOv7), the proposed object-to-spot rotation mainly aims to accurately predict the rotation angles. Hence, building a more rotation-specific deep model is of particular interest. For example, regressing the bounding boxes may be discarded and then instead replaced by object-centric points regression. This may guide the deep network to pay more attention to deriving the rotation knowledge rather than the bounding boxes.
\textbf{\textit{Finally}}, our attempt to bridge the gap between the synthetic and real data in terms of low-level properties such as texture, material, and smoothness level by using image smoothing and filtering techniques has shown a remarkable performance boost. However, it is worth trying cartonization technique on input real images which may provide better results and thus it is part of our future interest.

	\section{Conclusion}
In this paper, we leveraged the power of 3D graphics and computer vision techniques to tackle a real-world problem, that we propose object-to-spot rotation estimation which is of particular significance for intelligent surveillance systems, bike-sharing systems, and smart cities. We introduced a rotation estimator (OSRE) that estimates a parked bike rotation with respect to its parking area. By leveraging 3D graphics, we generated Synthetic Bike Rotation Dataset (SynthBRSet) with accurate bike rotation annotations. Next, we presented a first-of-its-type object rotation model that assesses bike parking. As we obtained promising results, we believe this work would be a significant starting point for further deep studies on rotation estimation with respect to other visual semantic elements in the input image.

	{\small
		\bibliographystyle{unsrt}
		\bibliography{ref}
	}

\pagebreak
\pagebreak

\section{Implementation Details}
In this section, we further provide more details of using Blender \cite{blender}, the generating script, and the generated synthetic dataset properties\footnote{All relevant data and codes to this paper are provided either as a supplementary file (with less than 250MB), or uploaded on cloud for large files (\ie, {\tt .blend}, assets, real data, and synthetic data) at {\tt \href{https://github.com/saghiralfasly/OSRE-Project}{https://github.com/saghiralfasly/OSRE-Project}}. We acknowledge Yolov7, Yolov5 and Yolor open sources as we built our rotation estimator on top of them:\\
	{\tt	\href{https://github.com/WongKinYiu/yolov7}{https://github.com/WongKinYiu/yolov7}}
	{\tt \href{https://github.com/WongKinYiu/yolor}{https://github.com/WongKinYiu/yolor}}
	{\tt \href{https://github.com/ultralytics/yolov5}{https://github.com/ultralytics/yolov5}}
}.

\subsection{Synthetic Dataset}
\noindent\textbf{Software and Hardware}: For generating the synthetic bike rotation dataset (SynthBRSet), we used the open source 3D graphics software Blender \cite{blender} $ 2.92.0 $ on $2$ $\times$ RTX $ 2080 $ Ti machine with Ubuntu 18.04.  

\noindent\textbf{Rendered Scene Augmentation}: The ease of automatically adjusting 3D virtual scenes and their visual objects, colors, lights, materials, textures, and camera-capturing angles leads to a highly visually reliable dataset. The strength implementation points of this dataset can be summarized in the following aspects: 
\textit{\textbf{Diversity}:} The dataset diversity, in terms of the output images, is enforced by enriching the rendered image with large variations of visual scenes, bike models, lights, textures, colors, buildings, and other 3D objects. We designed and collected several bike models, material textures, car models, external waste bins, poles, sidewalks umbrellas, trees, and road/pavement backgrounds. The diversity property can be further extended by adding more textures, scenes, 3D models, and objects.
\textit{\textbf{Balance}:} Generating visual data synthetically provides us full control to tackle the class imbalance. For example, the state when bikes are {\tt fallen} is with significantly less possibility than the state of {\tt rotated} bikes in the real world. Similarly, the possibility of seeing a rotated bike is less than the well-{\tt parked} state, and thus it leads to an imbalanced dataset when it is collected from real-world scenarios. Therefore, generating data synthetically using 3D graphics grants more reliability and generalization.
\textit{\textbf{Agnostic to Camera}:} We randomly render each image with various view angles by changing the camera locations in three dimensions. Specifically, we randomly locate the camera horizontally in the $x$ axis, vertically in the $z$ axis, and the depth in the $y$ axis. Although we organized our scene virtual environment in this axis structure, it can be adapted manually in a blender file, but in such a case the rotation axis structure of the entire environment should be adapted including the bike rotation readings.
\textit{\textbf{Size}.} The built 3D virtual generating framework can generate a large synthetic visual data with significantly less effort and less cost as compared to manual traditional methods which are much more tedious and laborious when it comes to collecting accurate rotations.   

\noindent\textbf{Running the Generating Code}: Since Blender is built on Python, it involves a scripting interface for editing and running python script directly. However, for faster image rendering and generating, we run the generating script from the command line in the background. Overall, the average time to render a single $360 \times 640$ image is $ 41 $ seconds.

\noindent\textbf{Rotation Reading:} The rotation reading $r$ is taken in degrees  $ 0^\circ \leq r \leq 360^\circ $ that ranges from $ 0^\circ $ which is well-{\tt parked} state to $ 360^\circ $ which is the complete rotation cycle representing the well-{\tt parked} state as well. However, these readings are then transformed  $ -180 \leq r \leq 180 $. In this way, we can make the rotation reading more intuitive for neural network learning. For example, the rotation angle $330^\circ$ is much larger in terms of its numeric measurable value, but it is close to the rotation angle $0^\circ$ with only $30^\circ$. Thus, we tend to read it as $-30^\circ$ which in turn leads to easier rotations input representation for neural networks. However, we visualize the output in two-nested circles, inner circle visualizes the bike rotation in the $y$ axis, whereas the rotation in the $z$ axis is visualized in the outer circle. 
\begin{figure*}
	\centering
	\includegraphics[trim=0 0 0 0, width=17.5cm,clip, keepaspectratio]{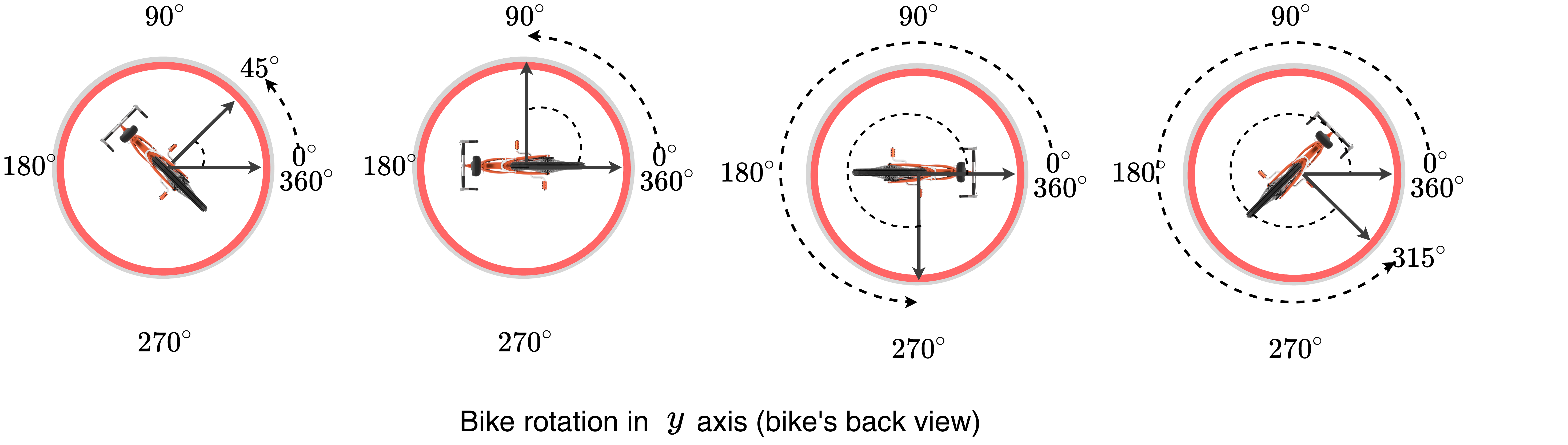}
	\caption{\textbf{Bike Rotation Visualization - back view.} The bike's predicted rotation in $y$ axis with respect to the parking area. This camera view (back view) is obtained when the camera is located in the lower location of its vertical path $z$. Note, we normalized this rotation into only $ 3 $ states: $0^\circ$ represents the standing well-parked state in $y$ axis, whereas $90^\circ$ and $-90^\circ$ represent the bike {\tt fallen} state in the right and the left side, respectively.}
	\label{fig:rotationY}
\end{figure*}

\begin{figure*}
	\centering
	\includegraphics[trim=0 0 0 0, width=17.5cm,clip, keepaspectratio]{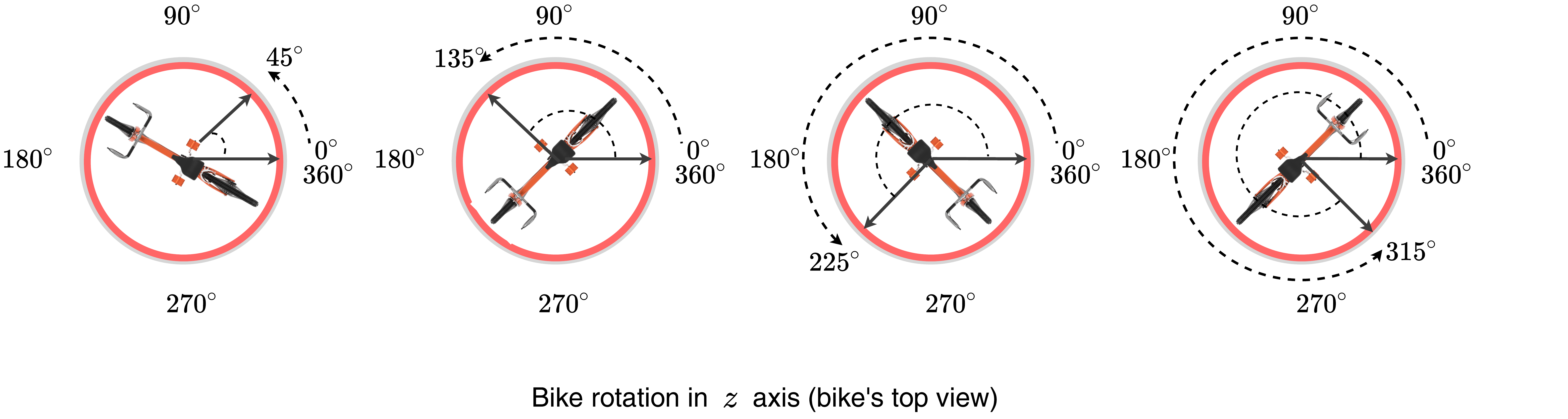}
	\caption{\textbf{Bike Rotation Visualization - top view.} The bike's predicted rotation in $z$ axis with respect to the parking area. This camera view (top view) is obtained when the camera is located in the middle location of its horizontal path $x$.}
	\label{fig:rotationZ}
\end{figure*}

\noindent\textbf{Rendered Images}: The image rendering is performed using GPU with Cycles render engine with $ 512 $ samples. We render the output images in $360 \times 640$ resolution and $jpg$ encoding of $\%98$ compression quality. Each rendered image contains a random number of bikes between ($ min, max $). However, these rendering parameters can be changed as we used {\tt config} file to allow updating them and others. for better visualizing the output in the demo and visual output samples, we rendered a set of images with $1080 \times 1920$ resolution.

\subsection{Real Data Collection}
\begin{figure}
	\centering
	\includegraphics[width=8.3cm,clip, keepaspectratio]{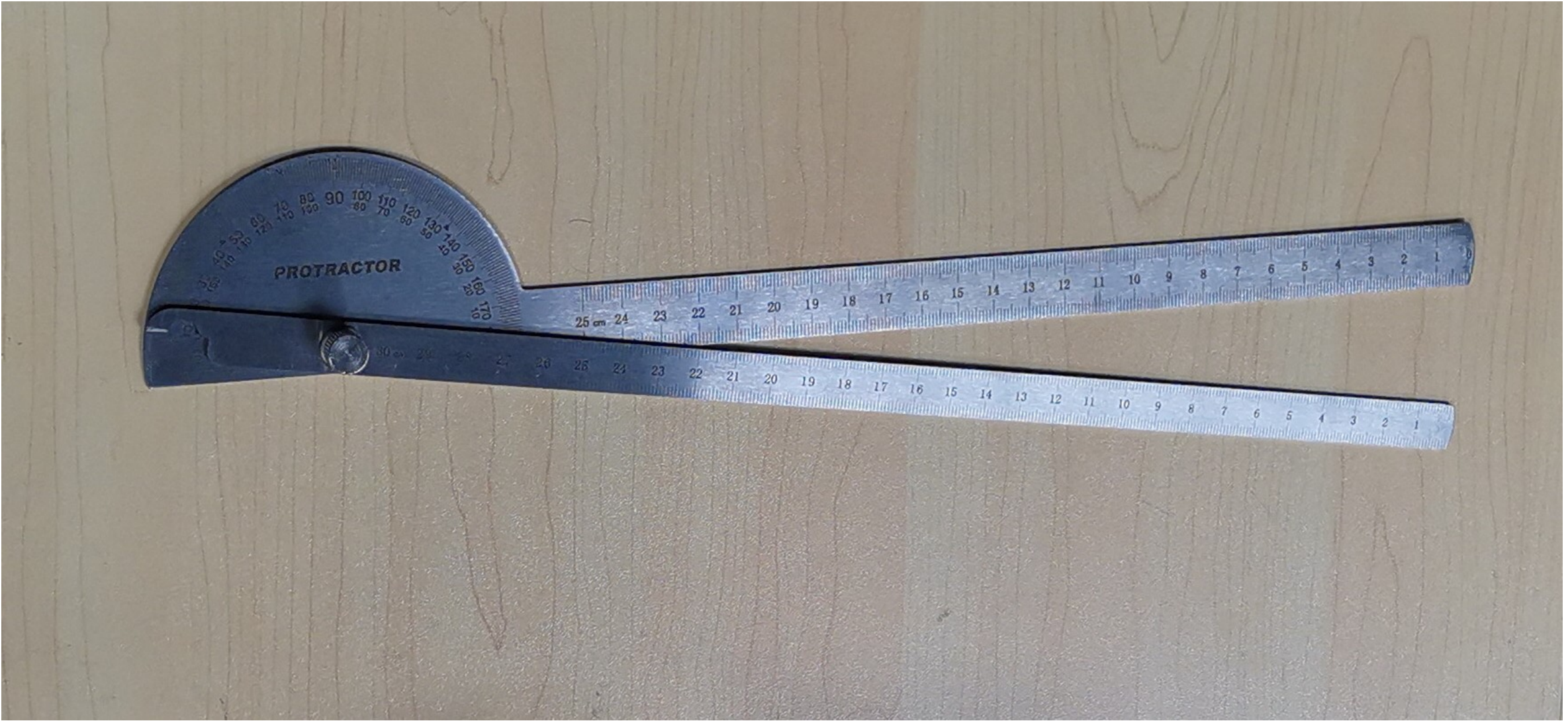}
	\caption{The used protractor for real-world bike rotation reading. First, we read the rotations in degrees. Then, we transform them into radians and get them normalized between $ 0 $ and $ 1 $. The base parking spot we collect the rotation with respect to is visually estimated by the overall real parking scene. For example, we have collected data from straight-aligned parking areas to the streets and ignored the rounded or curved parking areas.}
	\label{fig:protractor}
\end{figure}

Collecting real images for evaluating bike rotation with accurate rotation readings is difficult and laborious. However, it is important to collect some images for evaluating the proposed rotation estimator generalization as it is only trained on synthetic data. For more accurate rotation readings, we used the mobile camera to capture the images and protractor (see Fig. \ref{fig:protractor}) to measure the bike rotation with respect to its parking area. The parking area is not annotated as we do not annotate it in the synthetic data. This is to push the model during training to infer the parking spot in a semantic context rather than supervised the model to locate the parking spot area manually which may lead the trained model to the overfitting problem. 

\begin{figure*}
	\centering
	\includegraphics[trim=0 0 0 0, width=17.5cm,clip, keepaspectratio]{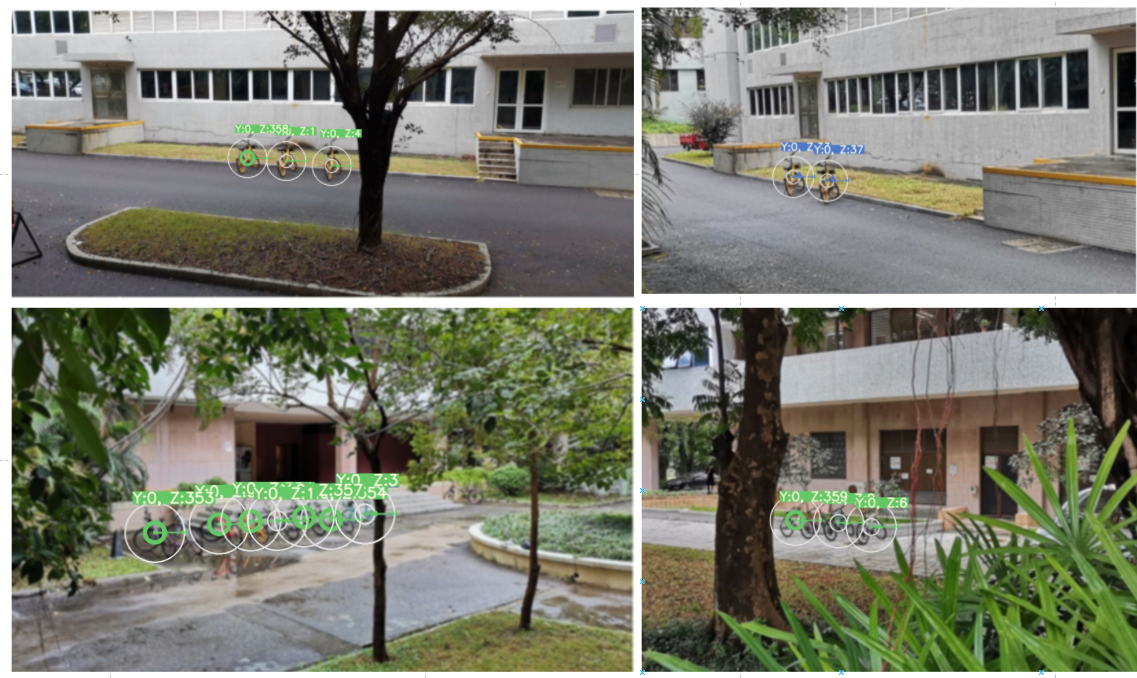}
	\caption{Output samples of the proposed bike rotation estimation method on real images with good estimation.}
	\label{fig:realSamples}
\end{figure*}

\begin{figure*}
	\centering
	\includegraphics[trim=0 0 0 0, width=17.5cm,clip, keepaspectratio]{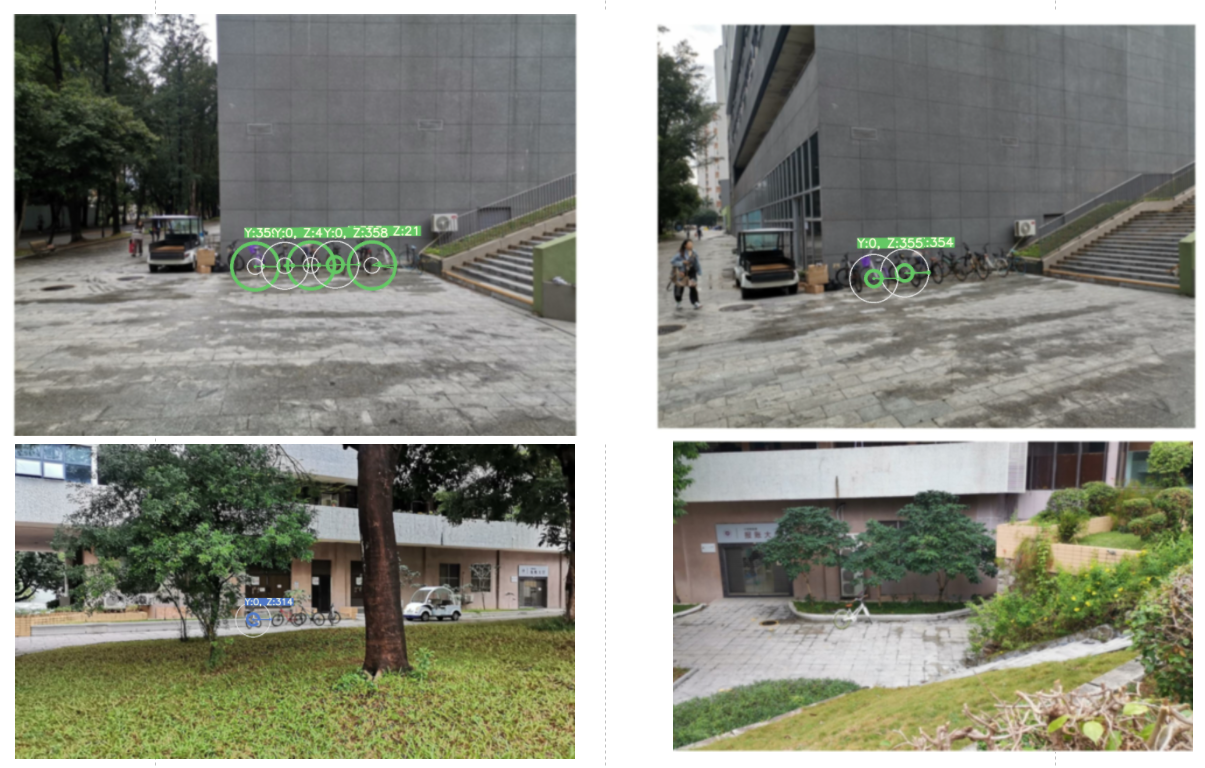}
	\caption{Output samples of the proposed bike rotation estimation method on real images with failed estimation.}
	\label{fig:samplesRealFaild}
\end{figure*}

\section{Output Rotation Visualization}
\begin{figure*}
	\centering
	\includegraphics[trim=0 0 0 0, width=17.5cm,clip, keepaspectratio]{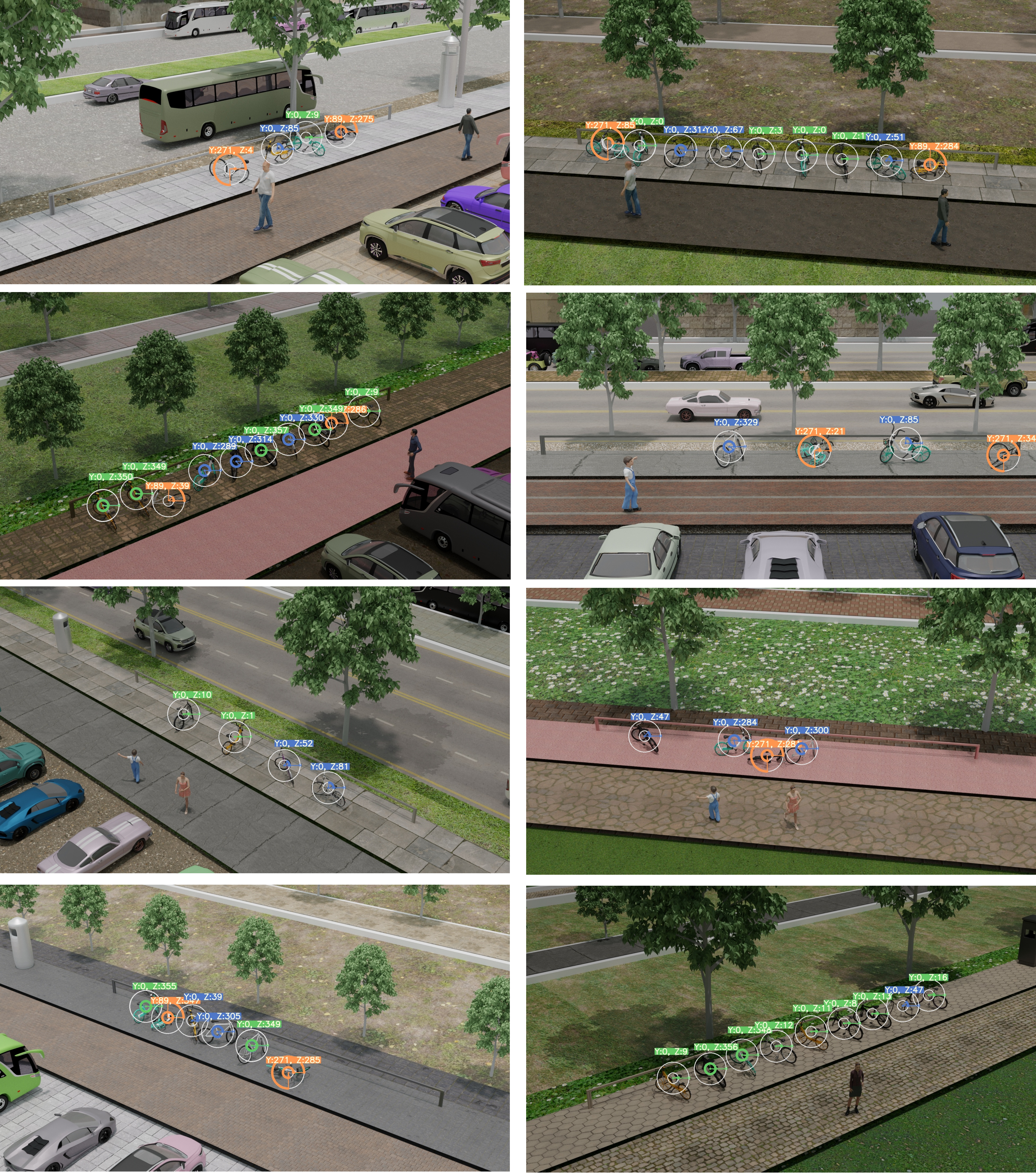}
	\caption{Output samples of the proposed bike rotation estimation method on synthetic images.}
	\label{fig:synthSamples2}
\end{figure*}

In the training phase, we shift the rotation readings from the range $[0^\circ,360^\circ]$ to  $[-180^\circ,180^\circ]$, then we transform them to radians and finally normalized them to the range  $[0,1]$. However, in the output rotation visualization, we perform the reverse transformation in order to get the real rotation in degrees. Fig. \ref{fig:rotationY} illustrates the rotation visualization in $y$ axis (bike's back view), whereas Fig.\ref{fig:rotationZ} illustrates the bike rotation in $z$ axis (bike's top view). Output visualizations in synthetic and real images are in Figs. \ref{fig:synthSamples2} and \ref{fig:realSamples}, respectively.

\section{Limitations}
Since our work is the first work that attempts to tackle the bike rotation estimation, there are several worth-mentioning weak points that can be improved and some parts can be further investigated. \textbf{\textit{First}}, the size of the collected real images is small. Thus, for better evaluation and further investigation of the proposed bike rotation estimation technique, it is recommended to collect more data with accurate rotation annotation in various real-world scenarios. \textbf{\textit{Second}}, the real image acquisition is performed using a mobile camera with various aspect ratios (rather than the regular CCTV aspect ratio 9:16). Hence, to transform the input image to the same aspect ratio used in the generated synthetic data and in most CCTVs, we have performed image padding while preserving their original aspect ratio. This is can be seen in the \href{https://drive.google.com/file/d/1pNxuI1dqqRJOOilJibBVtAZiy5oHZjJh/view?usp=share_link}{test set}.
\textbf{\textit{Third}}, our current synthetic data and trained bike rotation estimator are built for the scenario that the surveillance camera is located at any top/behind point of the bike parking area. Thus, as it is easy to adapt the proposed generating framework to create images with a front bike view can enrich the trained model generalization. 
\textbf{\textit{Finally}}, result visualization for the evaluation phase could be improved by visualizing the gap between the predictions and the ground truths.

\begin{figure*}
	\centering
	\includegraphics[trim=0 0 0 0, width=17.5cm,clip, keepaspectratio]{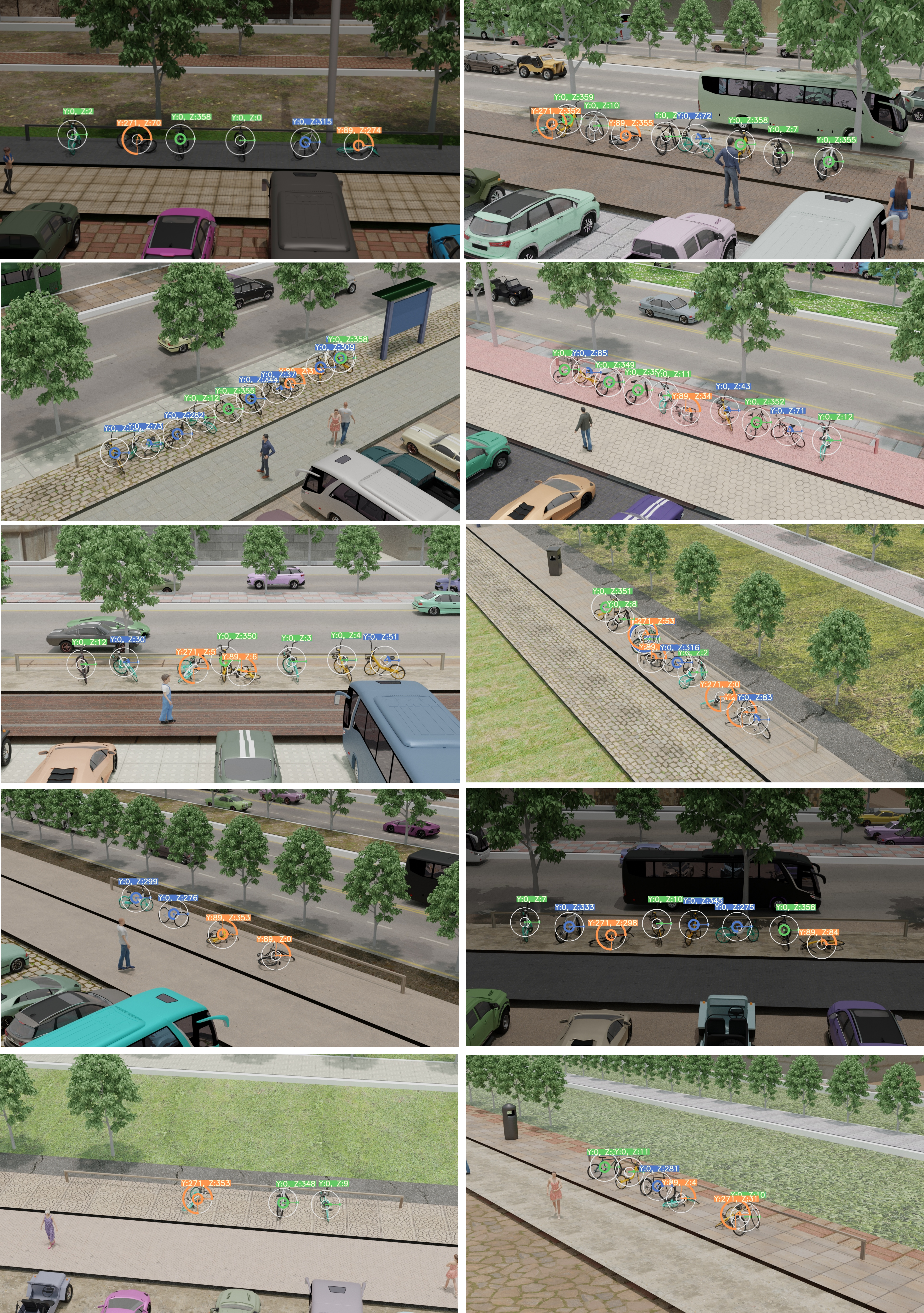}
	\label{fig:synthSamples1}
\end{figure*}

\end{document}